\documentclass[10pt,twocolumn,letterpaper]{article}

\usepackage[pagenumbers]{cvpr} %

\usepackage{graphicx}
\usepackage{amsmath}
\usepackage{amssymb}
\usepackage{booktabs}
\usepackage{arydshln}
\usepackage{algorithm}
\usepackage[noend]{algorithmic}
\usepackage{multirow}
\usepackage{amssymb}
\usepackage{gensymb}
\usepackage[dvipsnames]{xcolor}

\definecolor{orange}{rgb}{0.91, 0.52, 0.34}
\definecolor{green}{RGB}{118, 174, 87}
\definecolor{pink}{RGB}{195, 85, 132}

\usepackage[pagebackref,breaklinks,colorlinks]{hyperref}

\usepackage[capitalize]{cleveref}
\crefname{section}{Sec.}{Secs.}
\Crefname{section}{Section}{Sections}
\Crefname{table}{Table}{Tables}
\crefname{table}{Tab.}{Tabs.}

\newcommand{\name}{SPARTN}
\newcommand{\explainedname}{SPARTN (Synthetic Perturbations for Augmenting Robot Trajectories via NeRF)}
\newcommand{\dset}{\mathcal{D}}
\newcommand{\se}{\text{SE}(3)}
\newcommand{\E}{\mathbb{E}}
\newcommand{\tfm}[2]{{}^{#1}T^{#2}}
\newcommand{\Htfm}[2]{{}^{#1}H^{#2}}

\newcommand{\R}[1]{\textrm{Rot}\left[#1\right]}
\newcommand{\T}[1]{\textrm{Trans}\left[#1\right]}
\newcommand{\unif}{\mathcal{U}}

\def\cvprPaperID{8173} %
\def\confName{CVPR}
\def\confYear{2023}

\begin{document}

\title{NeRF in the Palm of Your Hand: \\Corrective Augmentation for Robotics via Novel-View Synthesis} %

\author{Allan Zhou\thanks{Equal contribution. Correspondence to \texttt{ayz@stanford.edu}.}\\
Stanford
\and
Moo Jin Kim$^*$\\
Stanford
\and
Lirui Wang\\
MIT CSAIL
\and
Pete Florence\\
Google
\and
Chelsea Finn\\
Stanford
}
\maketitle

\begin{abstract}
Expert demonstrations are a rich source of supervision for training visual robotic manipulation policies, but imitation learning methods often require either a large number of demonstrations or expensive online expert supervision to learn reactive closed-loop behaviors. In this work, we introduce \explainedname{}: a fully-offline data augmentation scheme for improving robot policies that use eye-in-hand cameras. Our approach leverages neural radiance fields (NeRFs) to synthetically inject corrective noise into visual demonstrations, using NeRFs to generate perturbed viewpoints while simultaneously calculating the corrective actions. This requires no additional expert supervision or environment interaction, and distills the geometric information in NeRFs into a real-time reactive RGB-only policy. In a simulated 6-DoF visual grasping benchmark, \name{} improves success rates by 2.8$\times$ over imitation learning without the corrective augmentations and even outperforms some methods that use online supervision. It additionally closes the gap between RGB-only and RGB-D success rates, eliminating the previous need for depth sensors. In real-world 6-DoF robotic grasping experiments from limited human demonstrations, our method improves absolute success rates by $22.5$\% on average, including objects that are traditionally challenging for depth-based methods. See video results at \url{https://bland.website/spartn}.
\end{abstract}

\setlength{\textfloatsep}{0.3cm}
\setlength{\floatsep}{0.3cm}

\section{Introduction}
\begin{figure}
\centering
\includegraphics[width=0.45\textwidth]{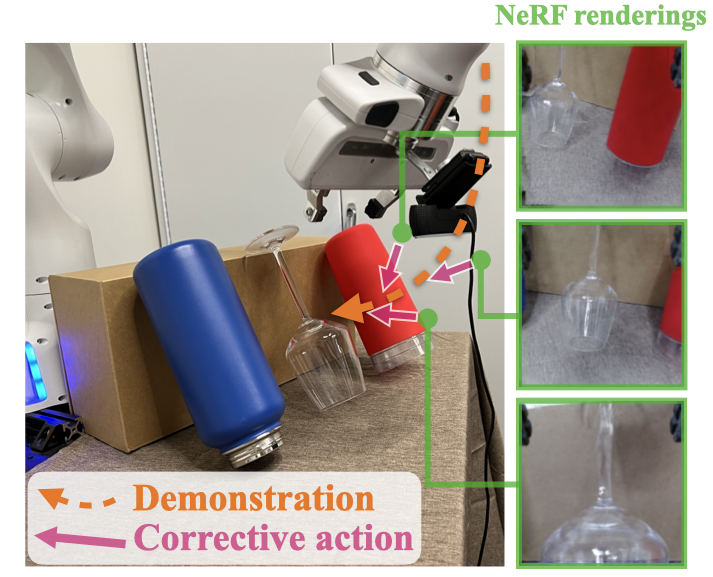}
\caption{\name{} is an offline data augmentation method for behavior cloning eye-in-hand visual policies. It simulates recovery in a \textcolor{orange}{demonstration} by using NeRFs to render high-fidelity \textcolor{green}{observations (right)} from noisy states, then generates \textcolor{pink}{corrective action} labels.}
\label{fig:front-figure}
\end{figure}

Object grasping is a central problem in vision-based control and is fundamental to many robotic manipulation problems. While there has been significant progress in top-down bin picking settings~\cite{dexnet2,qtopt}, 6-DoF grasping of arbitrary objects amidst clutter remains an open problem, and is especially challenging for shiny or reflective objects that are not visible to depth cameras. For example, the task of grasping a wine glass from the stem shown in Figure~\ref{fig:front-figure} requires precise 6-DoF control (using full 3D translation and 3D rotation of the gripper) and closed-loop perception of a transparent object. Traditional 6-DoF grasping pipelines \cite{dogar2011framework,wang2019manipulation} synthesize only one grasp pose and use a motion planner to generate a collision-free trajectory to reach the grasp~\cite{miller2004graspit,ten2017grasp,yan2018learning,mousavian20196}. However, the use of open-loop trajectory execution prevents the system from using perceptual feedback for reactive, precise grasping behavior. In this paper, we study how to learn closed-loop policies for 6-DoF object grasping from RGB images, which can be trained with imitation or reinforcement learning methods~\cite{gaddpg}.

Imitation learning from expert demonstrations is a simple and promising approach to this problem, but is known to suffer from compounding errors~\cite{ross2011reduction}. As a result, complex vision-based tasks can require online expert supervision~\cite{ross2011reduction,jang2022bc} or environment interaction~\cite{finn2016guided,rajeswaran2017learning}, both of which are expensive and time-consuming to collect. On the other hand, offline ``feedback augmentation'' methods~\cite{florence2019self,ke2021grasping} can be effective at combating compounding errors, but are severely limited in scope and thus far have not been applied to visual observations. Other recent works have found that using eye-in-hand cameras mounted on a robot's wrist can significantly improve the performance of visuomotor policies trained with imitation learning~\cite{hsuWristCamera,mandlekar2021matters,jangir2022look}, but still do not address the underlying issue of compounding errors. We develop an approach that helps address compounding errors to improve vision-based policies, while building on the success of eye-in-hand cameras.

To improve imitation learning for quasi-static tasks like grasping, we propose a simple yet effective offline data augmentation technique. For an eye-in-hand camera, the images in each demonstration trajectory form a collection of views of the demonstration scene, which we use to train neural radiance fields (NeRFs)~\cite{nerf} of each scene. Then, we can augment the demonstration data with corrective feedback by injecting noise into the camera poses along the demonstration and using the demonstration's NeRF to render observations from the new camera pose. Because the camera to end-effector transform is known, we can compute corrective action labels for the newly rendered observations by considering the action that would return the gripper to the expert trajectory. The augmented data can be combined with the original demonstrations to train a reactive, real-time policy. Since the NeRFs are trained on the original demonstrations, this method effectively ``distills'' the 3D information from each NeRF into the policy. 

The main contribution of this work is a NeRF-based data augmentation technique, called \explainedname{}, that improves behavior cloning for eye-in-hand visual grasping policies. By leveraging view-synthesis methods like NeRF, \name{} extends the idea of corrective feedback augmentation to the visual domain. The resulting approach can produce (i) reactive, (ii) real-time, and (iii) RGB-only policies for 6-DoF grasping. The data augmentation is fully offline and does not require additional effort from expert demonstrators nor online environment interactions. We evaluate \name{} on 6-DoF robotic grasping tasks both in simulation and in the real world. On a previously-proposed simulated 6-DoF grasping benchmark~\cite{gaddpg}, the augmentation from \name{} improves grasp success rates by $2.8\times$ compared to training without \name{}, and even outperforms some methods that use expensive online supervision. On eight challenging real-world grasping tasks with a Franka Emika Panda robot, \name{} improves the absolute average success rate by 22.5\%.

\section{Related Work}

\noindent \textbf{Robotic Grasping}. Grasping is a long-studied topic in robotics~\cite{shimoga1996robot}; see the multitude of survey articles for a complete review~\cite{bicchi2000robotic,bohg2013data,kleeberger2020survey}. Most data-driven grasping systems focus on learning how to predict some parameterized ``grasp'' (whether a full 6-DoF pose, 2-DoF table-top position, etc.), and leave intermediate motion generation to be open-loop, handled either through motion planners or simple heuristics,
e.g.~\cite{dexnet2,pinto2016supersizing,murali20206,NeuralDescriptorField,wang2019manipulation,fang2020graspnet}. Other works have trained \textit{closed-loop} grasping policies %
~\cite{morrison2018closing,qtopt,song2020grasping,gaddpg}, and bring all the benefits of closed-loop policies: including for example, the ability to avoid obstacles, to perform precise grasping without precise calibration, and to react to dynamic objects.
Additionally, grasping policies are often designed for top-down (2- or 3-DoF) grasping~\cite{lenz2015deep,pinto2016supersizing,qtopt}, while 6-DoF grasping typically requires depth or 3D information \cite{morrison2018closing,gaddpg,murali20206,song2020grasping}. In contrast, our method trains a reactive 6-DoF grasping policy with only RGB data. See Table~\ref{tab:grasp_rw} for a summary of the assumptions of the most related grasping works.

\begin{table}[t]
    \centering
    \small
\renewcommand{\arraystretch}{1.1}
\resizebox{1\linewidth}{!}{
\begin{tabular}{|c|c|c|c|c|}
\hline
{Method}&{No Depth} & {Full 6-DoF} & {Closed-Loop} \\ \hline
\cite{redmon2015real,dexnet2}& & &  \\
\cite{song2020grasping} & & \checkmark &  \checkmark     \\
\cite{ten2017grasp,mousavian20196,kerr2022evo,ichnowski2021dex} & &\checkmark & \\
\cite{qtopt,levine2018learning}& \checkmark & & \checkmark  \\
\cite{pinto2016supersizing}& \checkmark & &\\
\cite{morrison2018closing} & & & \checkmark \\
\cite{gaddpg,wang2022hierarchical} & & \checkmark  &\checkmark     \\
\name{} (ours) & \checkmark   & \checkmark & \checkmark \\
\hline
\end{tabular}}
\caption{Comparison of our approach with related grasping work. \name{} is the only approach to learn closed-loop 6-DoF grasping policies from only RGB inputs.}
\label{tab:grasp_rw}
\end{table}

\noindent \textbf{Imitation learning and data augmentation}. Behavior cloning is known to struggle with covariate shift: small errors cause imitation policies to fall slightly off of the data distribution and it is then difficult to correct the mistake back onto the data manifold. DAgger~\cite{ross2011reduction} and its variants~\cite{menda2019ensembledagger,kelly2019hg,hoque2021thriftydagger} mitigate this issue by obtaining expert corrections throughout training. Alternatively, DART~\cite{dart} injects noise during expert demonstration collection, which is especially effective with algorithmic experts but interferes with human demonstration collection. Previous works \cite{ke2021grasping,florence2019self} have injected noise into the low-dimensional system state after data collection (in a fully offline manner), but the visual observations are left out, limiting the helpfulness of noise injection. 
Our method can be seen as a visual, fully-offline version of noise injection that does not require perturbing the expert during demonstration collection, using NeRF to synthetically render perturbed states post-hoc. Unlike standard image augmentations for policy learning \cite{yarats2020image,laskin2020reinforcement}, our method uses NeRF to learn a 3D model of the demonstration scene, which enables us to generate high-fidelity novel views for data augmentation. In addition, while standard image augmentation approaches do not modify the action labels, we leverage hand-eye coordination %
to calculate corrective actions for augmented observations.

\noindent \textbf{NeRF for Robotics}. A number of recent works have investigated applications of NeRF and related methods in robotics, including
localization \cite{yen2021inerf}, navigation \cite{adamkiewicz2022vision}, dynamics modeling \cite{li20223d,driess2022learning,cleac2022differentiable}, reinforcement learning \cite{driess2022reinforcement}, and data generation for other learning-based methods \cite{yen2022nerf,ichnowski2021dex}.   NeRF-Supervision \cite{yen2022nerf}, for example, generates pixel-level correspondence to learn dense object descriptors, which are useful for manipulation tasks. %
For grasping, various methods have leveraged NeRF \cite{ichnowski2021dex,kerr2022evo,blukis2022neural} for open-loop grasp synthesis.
In contrast, our method uses NeRF offline to augment data for grasping and distills a reactive, real-time, RGB-only, closed-loop policy. %

\section{Methodology}

\begin{figure*}
\centering
\includegraphics[width=0.9\textwidth]{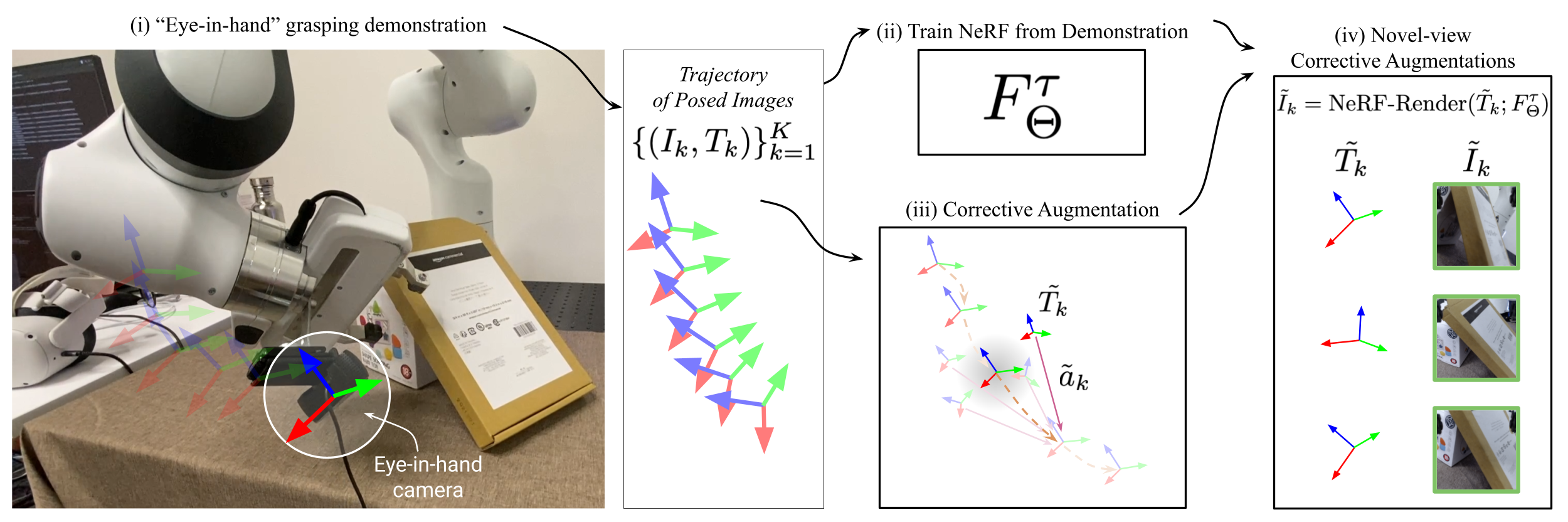}
\caption{An illustration of how \name{} creates augmentations from an original demonstration (in reality, this process is repeated for every available demonstration). \textbf{(i)}: The eye-in-hand demonstration contains posed images $\{(I_k, T_k)\}_{k=1}^K$. \textbf{(ii)}: We train a neural radiance field (NeRF) of the demonstration scene on the posed images. \textbf{(iii)}: We sample perturbations around each pose to \textbf{simulate} noise in the demonstration, and calculate the corrective action (in magenta) that would stabilize the trajectory. \textbf{(iv)}: We use the NeRF to render observations for the perturbed poses. The end result is augmented image-action pairs for improving behavior cloning.}
\label{fig:teaser}
\end{figure*}

We now introduce \name{}, which augments an eye-in-hand robot demonstration dataset using NeRF. We first review preliminaries, then describe a method overview, followed by details of training NeRFs and augmenting corrective behavior.

\subsection{Preliminaries}\label{subsec:preliminaries}

\noindent \textbf{Imitation learning.} In imitation learning, we assume access to a dataset of $N$ expert trajectories $\dset = \{\tau\}_{i=1}^{N}$, where each trajectory consists of a sequence of state-action pairs, $\tau = \{(s_k, a_k)\}_{k=0}^K$.  In purely \textit{offline} imitation learning, there exists no other primary data-collection assumptions other than this dataset $\dset$, i.e. no reward labels or online interactions.  A standard method for this offline setting is behavior cloning (BC),  which trains a policy $\pi_\theta$ to mimic the expert via supervised learning, by minimizing the objective $\mathcal{L}(\theta) = \E_{(s, a)\sim \dset}[\ell \big(\pi_{\theta}(s), a\big)]$, where $\ell$ is some loss function in the action space.

\noindent \textbf{NeRF.} Our method uses novel-view synthesis as a building block, for which we use Neural Radiance Fields~\cite{nerf}. 
For each scene, NeRF performs novel-view synthesis by training a scene-specific neural radiance field $F_{\Theta}$ from a ``training set'' of posed images $\{(I_k, T_k)\}$, where each $I_k \in \mathbb{R}^{w \times h \times 3}, T_k \in SE(3)$.
After $F_\Theta$ is trained, through volume rendering NeRF can render new views of a scene from any requested pose, which can be summarized as $I = \text{NeRF-Render}(T; F_{\Theta})$ -- in particular, this works best ``near'' the training set of poses.
Since we train many NeRFs (one per demonstration), we use an accelerated implementation of NeRF (Instant-NeRF\cite{mueller2022instant}).

\noindent \textbf{Corrective Noise Augmentation.} A simple method which has been shown to improve the robustness of behavior-cloned policies is to perform corrective noise augmentation \cite{florence2019self,ke2021grasping}. The idea is to take a nominal trajectory $\tau = \{(s_k, a_k)\}_{k=0}^K$ and create a noise-distributed corrective trajectory $\tilde{\tau} = \{(\tilde{s}_k, \tilde{a}_k )\}_{k=0}^K$, where each sampled state $\tilde{s}_k$ is a noisy version of the measured state, i.e. $\tilde{s}_k \sim s_k + \epsilon$ and $\epsilon$ is sampled noise. To perform corrective feedback augmentation, $\tilde{a}_k$ is chosen such that it will return the state to the nominal trajectory: given the true inverse dynamics $f^{-1}$ of the environment, then $\tilde{a}_k = f^{-1}({\tilde{s}_k, {s}_{k+1}})$, or compute the commanded control that can be tracked with a stabilizing controller to $s_{k+1}$. 
An easy way to parameterize this is by choosing the action space of the learned policy to be the input to a stabilizing controller, as in \cite{florence2019self,ke2021grasping}.  Note that it is also common in the imitation and reinforcement learning literature to apply noise to inputs, which is typically interpreted as a way to regularize the policy \cite{kostrikov2020image,laskin2020reinforcement}.  Meanwhile, in addition to potential regularization effects, the \textit{corrective} noise augmentation has been interpreted to specifically reduce compounding errors \cite{florence2019self,ke2021grasping}, but of course has limits if the scale of perturbations is too large or in highly non-smooth dynamics regimes.  A critical limitation of prior works using corrective noise augmentation is that they have not been applied to visual observations.

\subsection{Overview: Visual Corrective Augmentation} %

\begin{figure}[t]
\centering
\includegraphics[width=0.45\textwidth]{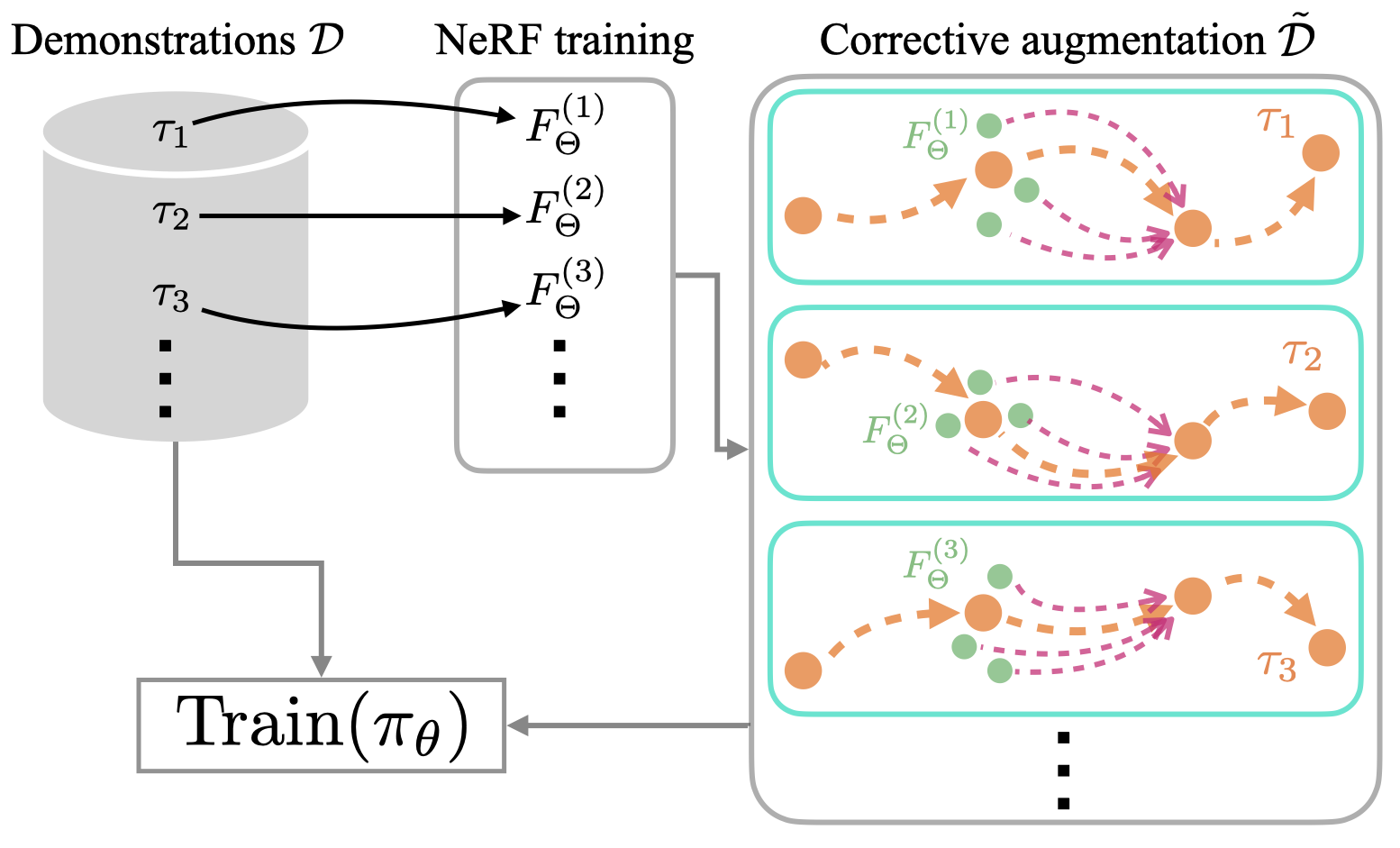}
\caption{An overview of the \name{} training process. A NeRF is trained for each of the original demonstrations in $\dset$. We use these NeRFs to generate visual corrective augmentations for each demonstration and collect them in $\tilde{\dset}$. The policy $\pi_\theta$ can be trained on $\dset$ and $\tilde{\dset}$ using standard behavior cloning methods.}
\label{fig:overview}
\end{figure}

\begin{algorithm}[t]
\caption{\name{} (Simplified overview)}
\label{alg:nerf-simple}
\begin{algorithmic}[1]
    \renewcommand{\algorithmicrequire}{\textbf{Input:}}
    \renewcommand{\algorithmicensure}{\textbf{Output:}}
    \renewcommand{\COMMENT}[1]{\textcolor{blue}{\# #1}}
    \REQUIRE $\dset = \{\tau\}_{i=1}^{N}$ - expert demonstrations
    \ENSURE Augmented transition dataset $\tilde{\dset}$
    \STATE $\tilde{\dset} \gets \{\}$
    \FOR{$\tau = \{(I_k, T_k, a_k) \}_{k=1}^K \in \dset$}
         \STATE $F^{\tau}_{\Theta} \gets \text{NeRF}(\{(I_k, T_k)\}_{k=1}^K)$ \COMMENT{train NeRF}
         \FOR{$k \in (0, 1, \ldots, K)$}
            \FOR{$i=1:N_{\text{aug}}$}
                \STATE $\varepsilon \gets \textrm{NoiseDist}(\se)$
                \STATE $\tilde{T}_k \gets \textrm{Perturb}(T_k, \varepsilon)$
                \COMMENT{perturb pose}
                \STATE $\tilde{a}_k \gets \textrm{CorrectAction}(T_k, \tilde{T}_k, a_k)$
                \STATE $\tilde{I}_k \gets \textrm{NeRF-Render}(\tilde{T}_k, F^{\tau}_{\Theta})$
                \STATE $\tilde{\dset} \gets \tilde{\dset} \cup \{(\tilde{I}_k, \tilde{T}_k, \tilde{a}_k)\}$
            \ENDFOR
        \ENDFOR
    \ENDFOR
\end{algorithmic}
\end{algorithm}

Consider the case where an agent receives partial visual observations $I$ instead of the full state of the world, and also receives direct proprioceptive state measurements, $s^{\text{robot}}$, as is commonly the case in robotics.  
In general, the corrective augmentation of Sec.~\ref{subsec:preliminaries} requires obtaining the visual observation $\tilde{I}_k$ for each noisy state $\tilde{s}_k$, which could be expensive or impossible in the actual environment. 

An insight of this work is that for \textit{eye-in-hand} robot policies (where the visual observation comes from a camera mounted on the wrist) in static scenes, we can readily generate novel visual observations using novel-view synthesis (i.e., NeRF) without further interactions in the environment. In this setting, a primary subset of the \textit{observations} are posed images $(I, T)$,  together with some \textit{actions} $a$.

The key intuition of our method can be grasped by considering how to perform visually corrective augmentation via NeRF, as illustrated in Figure~\ref{fig:teaser}. Noisy states and corrective actions $(\tilde{T}_k, \tilde{a}_k)$ can be generated for a trajectory of posed observations and actions $\tau = \{(I_k, T_k, a_k)\}_{k=0}^K$ (Sec.~\ref{subsec:preliminaries}).  The key pre-processing step is to train a trajectory-specific NeRF $F^{\tau}_{\Theta}$ for each demonstration $\tau$ (Sec.~\ref{subsec:nerf-training}).  These trajectory-specific NeRFs enable us to render observations $\tilde{I}_k$ for noisy states $\tilde{T}_k$, completing the augmentation process and resulting in visually corrective transitions $(\tilde{I}_k, \tilde{T}_k, \tilde{a}_k)$ (Sec.~\ref{subsec:visual-corrective-details}).  Algorithm~\ref{alg:nerf-simple} and Figure~\ref{fig:overview} overview this process.

\subsection{Training NeRFs from Robot Demonstrations}\label{subsec:nerf-training}

\name{} uses novel-view synthesis, in particular NeRF, to generate observations $\tilde{I}_k$ for noisy states without environment interaction. We train a NeRF $F^\tau_{\Theta}$ for each demonstration trajectory using the image observations $(I_1,\cdots,I_K) \in \tau$ as the training set of views. After training $F^{\tau}_{\Theta}$, we create observations $\tilde{I}_k$ for perturbed robot states $\tilde{T}_k$ by rendering the view from the perturbed camera pose using $F^{\tau}_{\Theta}$.

An important detail is that the end-effector reference frame used for control in demonstrations may differ from the reference frame for the camera itself, but through standard eye-in-hand calibration we can transform all visual observations used for training and augmenting the NeRFs into the cameras frame.  Given a transform $\tfm{\text{to}}{\text{from}}_k$ which transforms between two frames, we simply transform all NeRF-poses to the world frame: $\tfm{W}{C}_k = \tfm{W}{E}_k \tfm{E}{C}$, where $W$ is the world frame, $E$ is the end-effector frame which changes at each step $k$, and $C$ is the camera (NeRF) frame statically linked to the end-effector frame, and $^ET^C$ is acquired through hand-eye calibration.

\noindent\textbf{COLMAP camera poses.} Real-world calibration error means that our camera-to-world transforms $\{\tfm{W}{C}_k\}_{k=1}^K$ are noisy and we obtain higher-quality NeRFs by using camera poses estimated by COLMAP\cite{schoenberger2016sfm,schoenberger2016mvs}. Up to noise and a scale factor $\beta$, the only difference from the world frame camera transforms is that COLMAP uses an arbitrary reference frame $V\neq W$. We denote the COLMAP outputs $\{\Htfm{V}{C}_k\}_{k=1}^K$, using $H$ instead of $T$ because of the difference in scale. We now introduce notation to separate the rotation and translation components of a transform:
\begin{equation}
    \tfm{a}{b} := \left(\R{\tfm{a}{b}}, \T{\tfm{a}{b}}\right)
\end{equation}
Since we train the NeRFs on COLMAP's camera poses, we must convert perturbed camera poses $\tfm{W}{\tilde{C}}_k$ to COLMAP's frame in order to render the observations. In other words, we must call $\text{NeRF-Render}(\Htfm{V}{\tilde{C}}_k, F_{\Theta}^\tau)$, where:
\begin{align}
    \label{eq:colmap}
    \Htfm{V}{\tilde{C}}_k &= \left(\R{\tfm{V}{\tilde{C}}_k}, \textcolor{blue}{\beta} \ \T{\tfm{V}{\tilde{C}}_k}\right) \\
    \tfm{V}{\tilde{C}}_k &= \textcolor{blue}{\tfm{V}{W}} \ \tfm{W}{\tilde{C}}_k.
\end{align}
Both $\beta$ and $\tfm{V}{W}$ can be estimated from the pairs $\{(\tfm{W}{C}_k, \Htfm{V}{C}_k)\}_{k=1}^K$, as described in Appendix~\ref{appendix:colmap-nerf}.

\noindent \textbf{Static Scene Assumption.} An additional consideration for robotic manipulation is that the standard NeRF formulation assumes a static scene, while manipulation tasks such as grasping will usually move objects in the scene. To address this, we apply the NeRF training and augmentation process to only the subsets of each demonstration trajectory where no robot-object interaction occurs, instead of all timesteps. For grasping, a simple and effective heuristic is to only apply \name{} to the portions of each demonstration \textit{before} the expert closes the gripper.

\begin{figure}
\centering
\includegraphics[width=0.47\textwidth]{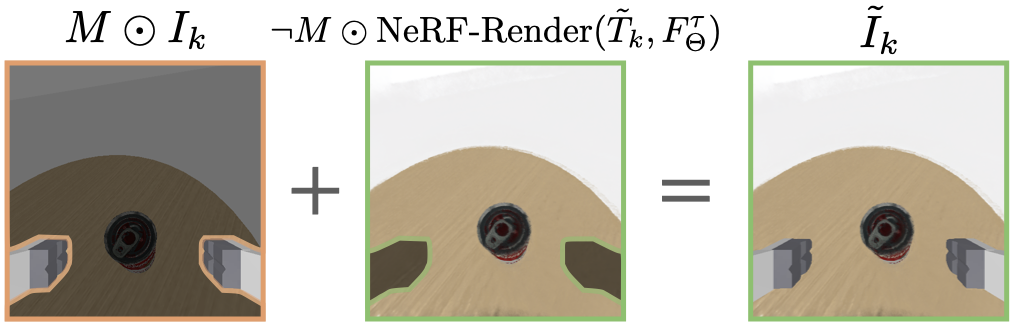}
\caption{An illustration of how the gripper is inserted into the result of the NeRF rendering process. Gray regions indicate pixels being masked out by the binary gripper mask $M\in\{0,1\}^{w \times h}$, which denotes the pixels where the gripper is located in all frames.}
\label{fig:masking}
\end{figure}

\noindent \textbf{Masking Out the Robot Gripper.} The robot's gripper is often within the view of an eye-in-hand camera, which breaks the static scene assumption. %
To address this, we leverage that the gripper is in the same location in each image assuming the camera is rigidly mounted and the gripper is open. Further, since NeRF is trained per-ray, we can simply mask pixels from training images. We construct a single binary mask $M\in\{0,1\}^{w\times h}$, where a $1$ indicates gripper pixels to mask out. Figure~\ref{fig:masking} shows how we use the same mask to splice the gripper back into each NeRF rendering output, $\tilde{I}_k \leftarrow \neg M  \odot \tilde{I}_k + M \odot I_k $, where $\odot$ is element-wise multiplication broadcasted across the color channels.

\noindent \textbf{NeRF Quality.} Even when the training views from a demonstration are suboptimal for NeRF training, \name{} benefits from the fact that our augmentation is local and the perturbations $\varepsilon$ are typically small, so we only need the NeRF to generalize in a small region around the demonstration trajectory. As we will verify in the experiments, \name{} can be effective even with a limited number of training views compared to other NeRF applications.

\subsection{NeRFing Corrective Noise Augmentation}\label{subsec:visual-corrective-details}

Given a visual augmentation model (Sec.~\ref{subsec:nerf-training}), we can adapt methods for corrective augmentation (Sec.~\ref{subsec:preliminaries}) into the visual domain. Our goal is to create noise-distributed corrective transitions $\{(\tilde{I}_k, \tilde{T}_k, \tilde{a}_k)\}$.  First, we describe this simply in the global frame. In order to sample from the noise-distributed corrective trajectory, one can first apply noise to the measured end-effector pose, $\tilde{T}_k := T_k \varepsilon$, where $\varepsilon \sim \text{NoiseDist}(SE(3))$ is a randomly sampled rotation and translation.
The high-fidelity perturbed image $\tilde{I}_k$ corresponding to $\tilde{T}_k$ can then be rendered using the trajectory-specific NeRF $F^{\tau}_{\Theta}$, without requiring access to the actual environment. 
For the actions, the simplest case is when they are inputs to a global-frame stabilizing Cartesian controller controller \cite{khatib1987unified}, in which case $\tilde{a}_k=a_k=\hat{T}_k$ will provide stabilization to the nominal trajectory, where $\hat{T}_k$ is the \textit{desired} pose sent to the lower-level controller.

\noindent \textbf{Corrective Relative Actions.} As is common in prior works, we observe better performance by parameterizing the learned policy as a \textit{relative} rather than global action.  Consider the as-discussed global-frame version, with (1) \textit{observations} as a global-frame measured SE(3) end-effector pose $\tfm{W}{E}_k$, where $W$ refers to world-frame, and $E$ to the end-effector frame at timestep $k$, and (2) \textit{action} as a global-frame desired SE(3) end-effector pose $\tfm{W}{\hat{E}}_k$. %
To switch to a relative action space, we adjust the action to
\begin{equation}
a_k=\tfm{E}{\hat{E}}_k = (\tfm{W}{E}_k)^{-1} \ \tfm{W}{\hat{E}}_k = \ \tfm{E}{W}_k \tfm{W}{\hat{E}}_k.
\end{equation}
To additionally formulate the corrective noise augmentation in the relative frame, we consider the SE(3)-noise $\varepsilon$ as transforming from the noisy end-effector frame to the measured end-effector frame, i.e. $\tfm{E}{\tilde{E}} := \varepsilon$.  This accordingly adjusts the observation as $\tfm{W}{\tilde{E}}_k = \ \tfm{W}{E}_k \tfm{E}{\tilde{E}}_k = \ \tfm{W}{E}_k \varepsilon$
and the \textit{relative} action as:
\begin{equation}
    \tilde{a}_k=\tfm{\tilde{E}}{\hat{E}}_k
    = (\tfm{W}{E}_k \ \tfm{E}{\tilde{E}}_k)^{-1} \  \tfm{W}{\hat{E}}_k
    = \varepsilon^{-1} \ a_k.
\end{equation}
Concisely, this amounts to post-pending $\varepsilon$ to the measured pose, and pre-prending $\varepsilon^{-1}$ to the un-noised relative action.

\noindent \textbf{Non-$\se$ actions.} Thus far we have assumed that the actions $a\in \se$ only command desired pose transforms. In practice, the action space may contain \textit{additional} dimensions, for example to open and close the gripper itself. The \name{} augmentation does not concern these aspects of the action, so the corrective actions simply copy the original action for non-$\se$ dimensions.

\noindent \textbf{Summary.} Figure~\ref{fig:teaser} summarizes our augmentation procedure: given a demonstration dataset $\dset$, we first train all the neural radiance fields $\{F^\tau_{\Theta}\}_{\tau \in \dset}$ and save the weights to disk. We then augment each transition in the original dataset with $N_{\text{aug}}$ noisy corrective transitions produced using the process we described above, and save these transitions into an augmented dataset $\tilde{\dset}$. Appendix Algorithm~\ref{alg:nerf} describes the precise augmentation procedure in detail.

After the augmented dataset has been created, it can simply be combined with the original dataset to augment BC training. Various sampling strategies are possible, but in our experiments we simply construct mini-batches for BC training by sampling from the original and augmented datasets with equal probability.

\section{Simulation Experiments}
\begin{figure}
\centering
\includegraphics[width=0.45\textwidth]{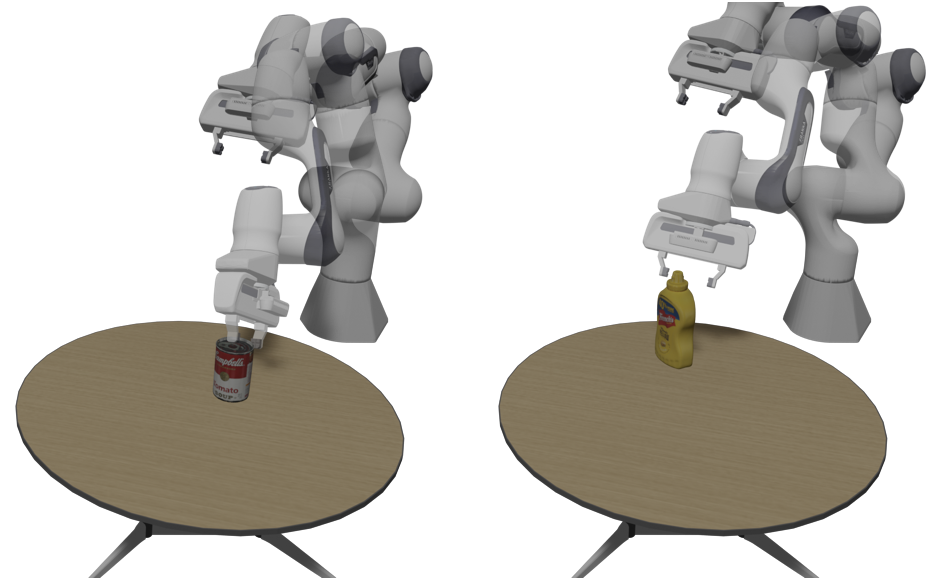}
\caption{Example tasks in the simulated 6-DoF grasping benchmark, first introduced in \cite{gaddpg}. There are $\sim 1{,}500$ ShapeNet objects in training demonstrations, and held out YCB objects for evaluation. A camera mounted to the Franka Panda robot's arm provides observations, and the policy controls the 6-DoF end-effector pose.}
\end{figure}

We evaluate \name{} and related approaches in the simulated 6-DoF grasping benchmark first introduced in \cite{gaddpg}, which features a simulated Franka Emika Panda arm with a parallel-jaw gripper. Objects are placed on a table and must be lifted above a height threshold in a successful grasp. Policies receive either RGB, RGBD, or point cloud observations from a wrist camera, and control the gripper by commanding relative pose changes in 6-DoF end-effector space.

\subsection{Data Collection and Evaluation Protocol} We follow the training and evaluation procedure from~\cite{gaddpg}. The training dataset includes $2{,}500$ demonstrations of grasping $1{,}500$ ShapeNet~\cite{chang2015shapenet} objects. Demonstrations are up to $20$ timesteps and are generated by trajectory optimization to precomputed grasps from the ACRONYM dataset~\cite{eppner2020acronym}. Policies are evaluated on grasping \textit{held-out} objects from the YCB~\cite{calli2015benchmarking} or ShapeNet datasets. Though it is more ``out of distribution'' relative to the training objects, the YCB evaluation is more realistic for tabletop grasping scenarios (ShapeNet includes a wide variety of objects, e.g. airplanes and bicycles). Each held-out object is evaluated ten times, each with a different initial robot configuration and object's initial pose.

\subsection{Comparisons}
\begin{table}
\begin{center}
\resizebox{0.99\columnwidth}{!}{%
\begin{tabular}{ c c c l l}
\toprule
Supervision & Method & Input & YCB SR(\%) & SN SR(\%)\\
\hline
\multirow{6}{*}{Offline} & BC & RGB & $28.9 \pm 2.4$ & $57.4 \pm 0.2$\\
\cdashline{2-5}
& \multirow{3}{*}{$\textrm{DART}^{\dagger}$} & RGB & $51.2 \pm 3.2$ & $57.8 \pm 2.2$\\
& & RGBD & $\mathit{46.3}$ & $\mathit{45.3}$\\
& & Point cloud & $\mathit{65.6}$ & $\mathit{73.6}$ \\
\cdashline{2-5}
& HA & RGB & $29.7\pm 2.4$ & $57.5 \pm 0.8$\\
\cdashline{2-5}
& \textbf{\name{} (ours)} & RGB & $\mathbf{74.7} \pm 2.4$ & $\mathbf{66.9} \pm 1.6$\\
\hline
\multirow{4}{*}{\textcolor{gray}{Online*}} & \multirow{3}{*}{\textcolor{gray}{DAgger}} & \textcolor{gray}{RGB} & \textcolor{gray}{$\mathit{52.3}$} & \textcolor{gray}{$\mathit{52.1}$}\\
& & \textcolor{gray}{RGBD} & \textcolor{gray}{$\mathit{67.1}$} & \textcolor{gray}{$\mathit{60.4}$}\\
& & \textcolor{gray}{Point cloud} & \textcolor{gray}{$\mathit{77.2}$} & \textcolor{gray}{$\mathit{75.8}$}\\
\cdashline{2-5}
& \textcolor{gray}{GA-DDPG} & \textcolor{gray}{Point cloud} & \textcolor{gray}{$\mathit{88.2}$} & \textcolor{gray}{$\mathit{91.3}$} \\
\bottomrule
\end{tabular}
}
\end{center}
\vspace{-1em}
\caption{Grasping success rates (SR) on \textit{held-out objects} from YCB~\cite{calli2015benchmarking} or ShapeNet (SN)~\cite{chang2015shapenet} in a simulated 6-DoF grasping benchmark \cite{gaddpg}. We \textbf{bold} the best offline RGB-only results, though we include online and non-RGB methods for comparison.
\textcolor{gray}{Online*} requires additional environment interactions, while Offline only uses demonstration data. $\textrm{DART}^{\dagger}$ is offline but requires a special demonstration collection setup.  \name{} outperforms other offline RGB-only methods, while RL (GA-DDPG) performs best overall while requiring millions of interactions. We calculate average success rates and standard error over $4$ random seeds. \textit{Italicized} success rates were reported in prior work\protect\footnotemark~\cite{gaddpg}.}
\label{table:main-results}
\end{table}
\footnotetext{Note that the ``BC'' results in \cite{gaddpg} actually use DART.}

We compare \name{} against other approaches ranging from simple behavior cloning to online-supervised imitation and reinforcement learning:

\noindent
\textbf{Behavior Cloning (BC)}: Trains policies via supervised learning on the demonstration dataset.

\noindent
\textbf{DART}~\cite{dart}: Introduces a modified demonstration setup where a continuous Gaussian noise is injected into the robot's state \textit{during} the expert demonstration, then trains policies on the modified demonstrations with BC.

\noindent
\textbf{Homography Augmentation (HA)}: A simplification of \name{} where perturbations can be 3D rotations, but not translations, of the camera. For pure rotation, we can calculate the homography transform for rendering the rotated view without NeRF. Augmented actions are computed similarly to \name{}.

\noindent
\textbf{DAgger}~\cite{ross2011reduction}: An online-supervised method where a policy is first trained using BC on offline demos, and then the expert provides action labels for states from the policy's rollouts throughout further policy training.

\noindent
\textbf{GA-DDPG}~\cite{gaddpg}: Jointly trains policies via BC and finetunes them with reinforcement learning (DDPG~\cite{silver2014deterministic}).

Of the methods considered, DAgger and GA-DDPG require online environment interaction and supervision from an expert or reward function, respectively. The other methods, including \name{}, train only on pre-collected demonstration datasets. Since the DART data collection setup noisily perturbs the robot as the expert demonstrates, DART
can interfere with human expert demonstration, though this challenge does not arise here with our simulated expert.

\begin{table}
\begin{center}
\resizebox{0.6\columnwidth}{!}{
\begin{tabular}{ c c l }
\toprule
 Method & Image aug. & YCB SR(\%) \\ 
 \hline
 \multirow{2}{*}{BC} & Without & $25.3 \pm 1.6$ \\
    & With & $28.9 \pm 2.4$ \\
\hdashline
 \multirow{2}{*}{DART} & Without & $51.2 \pm 3.2$ \\
    & With & $48.6 \pm 2.8$ \\
\hdashline
 \multirow{2}{*}{HA} & Without & $23.9 \pm 4.0$ \\
    & With & $29.7 \pm 2.4$ \\
\hdashline
 \multirow{2}{*}{\name{}} & Without & $68.3 \pm 3.6$ \\
    & With & $\mathbf{74.7} \pm 2.4$ \\
\bottomrule
\end{tabular}
}
\end{center}
\vspace{-1em}
\caption{Ablating the effect of standard image augmentation on each method for RGB policies. Average success rates and standard errors are calculated over four seeds.}
\label{table:augmentation}
\end{table}

\subsection{Training Details}
We follow the RGB architectural and training hyperparameters of \cite{gaddpg}, with policies consisting of a ResNet-18~\cite{he2015deep} image encoder followed by an MLP. We apply random crop and color jitter to the training images. We re-use the same training BC hyperparameters for \name{}. Appendix~\ref{appendix:sim-training} describes the complete training details.

For \name{}, we create $N_{aug}=100$ augmented transitions from each original transition and save the augmented dataset to disk before BC training.
To sample the perturbations $\varepsilon \sim \text{NoiseDist}(\se)$, we parameterize the rotations in terms of Euler angles $(\phi,\theta,\varphi)$ and uniformly sample both rotation and translation parameters:
\begin{align}
\label{eq:noisedist}
(\phi,\theta,\varphi), (t_x, t_y, t_z) &\sim \unif(-\alpha, \alpha),  \unif(-\beta, \beta) \\
\varepsilon &:= \left(R(\phi,\theta,\varphi), (t_x, t_y, t_z)\right)
\end{align}
In simulation, we set $\alpha=0.2\;\text{radians}$ and $\beta=3\;\text{mm}$. Following the DART hyperparameters in \cite{gaddpg}, we only augment timesteps $5-13$ of each demonstration. Appendix~\ref{sec:perturbed-samples} contains samples of \name{}'s perturbed observations rendered via NeRF.
\subsection{Results}
Table~\ref{table:main-results} shows grasping success rates on held-out objects from either the YCB or ShapeNet (SN) datasets. \name{} significantly outperforms the other two offline-supervised approaches for RGB inputs. Since DART injects noise during expert collection, the amount of augmentation is limited by the number of demonstrations the expert can collect. Meanwhile, \name{} can cheaply generate an arbitrary number of augmented examples on top of the existing demonstration dataset, leading to more data diversity without any additional effort from the expert. See the supplementary material for videos of rollouts for \name{} and BC policies, with success and failure cases.

Naive BC performs relatively well on ShapeNet evaluation, likely because the evaluation set is ``in-distribution''. Correspondingly, corrective methods (DART, \name{}) achieve smaller improvements over BC on ShapeNet and larger improvements on YCB.

On YCB, \name{}'s performance is closer to RL (GA-DDPG) than to other offline methods. It actually outperforms DAgger for RGB policies, perhaps because DAgger's data distribution changes online presenting a more challenging learning problem. Notably, on YCB \name{} outperforms DART with point clouds and is comparable to DAgger with point clouds. Since \name{} itself only requires RGB images during both NeRF training and policy training, it significantly closes the gap between the performance of RGB-only and the best depth-based methods. Since consumer depth cameras can struggle with common thin and reflective items~\cite{yen2022nerf}, improving RGB-only grasping can enable more robust grapsing of such objects.

\noindent \textbf{Effect of Image Augmentations.} We try ablating the effect of standard image augmentations (random crop and color jitter) for the BC, DART, HA, and \name{} methods.
Table~\ref{table:augmentation} shows that the image augmentations have a small effect on the performance of most methods, relative to the larger effect of \name{}'s corrective augmentations.

\section{Real-World Experiments}

\begin{figure}
\centering
\includegraphics[width=0.47\textwidth]{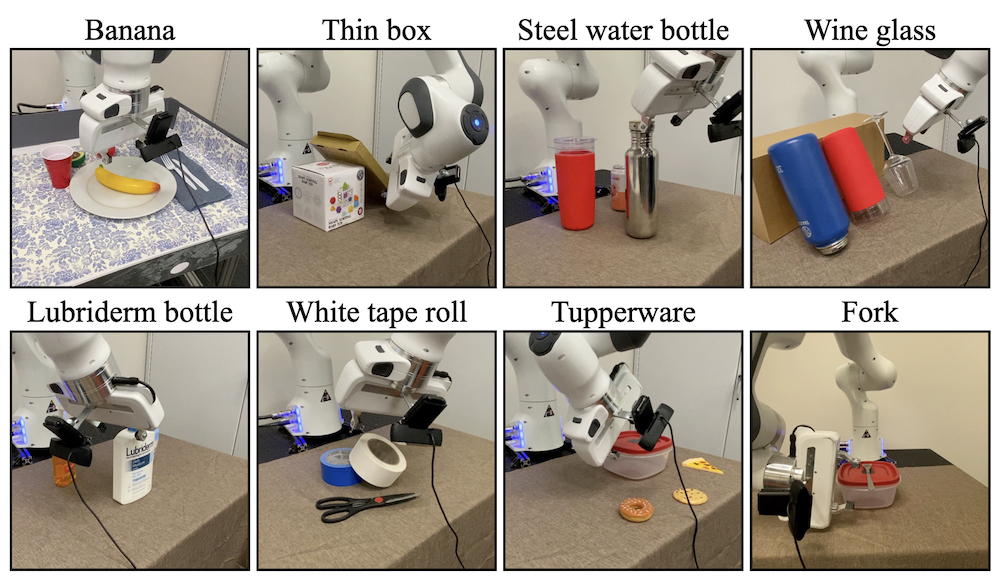}
\vspace{-0.3cm}
\caption{Real-world grasping environments. In each environment, the task is to grasp the labeled target object in a particular way. 
The target objects have various geometric shapes
and exhibit a diverse range of characteristics, including reflectiveness, transparency, and radial symmetry.}
\label{fig:environments}
\end{figure}

The simulation benchmark results show that \name{} can improve grasping generalization in imitation learning without online supervision. Here, we verify that \name{} can enable real-world robotic grasping of challenging objects from limited human demonstration. See the website in the supplement for video results.

\begin{table}
\begin{center}
\resizebox{0.48\textwidth}{!}{
\begin{tabular}{ c c c c }
\toprule
 Target Object & \# Demos & BC SR(\%) & \textbf{\name{} SR(\%)} \\ 
 \hline
 Banana & $14$ & $55$ & $\mathbf{75}$ \\
 Thin box & $20$ & $35$ & $\mathbf{65}$ \\
 Steel water bottle & $15$ & $20$ & $\mathbf{40}$ \\
 Wine glass & $25$ & $70$ & $\mathbf{90}$ \\
 Lubriderm bottle & $17$ & $25$ & $\mathbf{60}$ \\
 White tape roll & $15$ & $40$ & $\mathbf{45}$ \\
 Tupperware & $20$ & $40$ & $\mathbf{75}$ \\
 Fork & $20$ & $25$ & $\mathbf{40}$ \\
\hdashline
 Average & \texttt{--} & $38.75$ & $\mathbf{61.25}$ \\
\bottomrule
\end{tabular}
}
\end{center}
\vspace{-1em}
\caption{Success rates (SR) of behavior cloning (BC) and \name{} 6-DoF grasping policies on a suite of eight real-world target objects. \name{} outperforms BC in every environment, achieving an average absolute performance boost of $22.5\%$. Each success rate is computed over $20$ trials.}
\label{table:real-world-results}
\end{table}

\subsection{Experimental details}
\noindent\textbf{Robot setup.} The robotic manipulator is a Franka Emika Panda robot arm with a wrist-mounted consumer grade webcam. Policies take images of size $256\times256$ as input and output a 6-DoF action representing the desired change in end-effector position and orientation, as well as a binary open/close gripper action. We use a Cartesian impedance controller to command the pose changes at a frequency of 4 Hz. We task the robot to grasp a target object in eight different environments depicted in Figure~\ref{fig:environments}. The target objects include natural shapes that are common in the real world and exhibit a diverse range of attributes, such as reflectiveness, transparency, and radial symmetry.

\noindent\textbf{Comparisons and Evaluation.} In each environment, we collect a small number of expert grasping demonstrations with a virtual reality controller. Because DART is difficult to use with human demonstration collection, we compare \name{} to a vanilla BC policy trained on the same set of demonstrations. Policies are evaluated with the same objects seen in the demonstrations. Initial object and robot configurations are randomized during both data collection and evaluation. %

\noindent\textbf{Training.} To improve NeRF quality for \name{}, we program the robot to collect a few images of the scene from a fixed set of poses before the start of each demonstration. This automatic process is only used to improve COLMAP's pose estimation and the subsequent NeRF training.
For \name{}, we generate $N_{aug}=50$ augmented transitions from each original transition in the demonstrations. We sample perturbations $\epsilon$ according to Eq.~\ref{eq:noisedist} with $\alpha=0.05\;\text{radians}$ and $\beta=0.4\;\text{mm}$. Appendix~\ref{appendix:colmap-nerf} describes COLMAP and NeRF training in more detail, and Appendix~\ref{sec:perturbed-samples} shows sample \name{} observations rendered by NeRF. Aside from using the augmented dataset, \name{} policies are trained using the same BC architecture and training hyperparameters described in Appendix~\ref{sec:real-world-policy-training}.

\subsection{Results}
Table~\ref{table:real-world-results} shows grasping success rates in the eight real-world environments. Quantitatively, \name{} policies outperform the baseline BC policies across the board, on average achieving an absolute $22.5\%$ increase in success rate.

\begin{figure}
\centering
\includegraphics[width=0.48\textwidth]{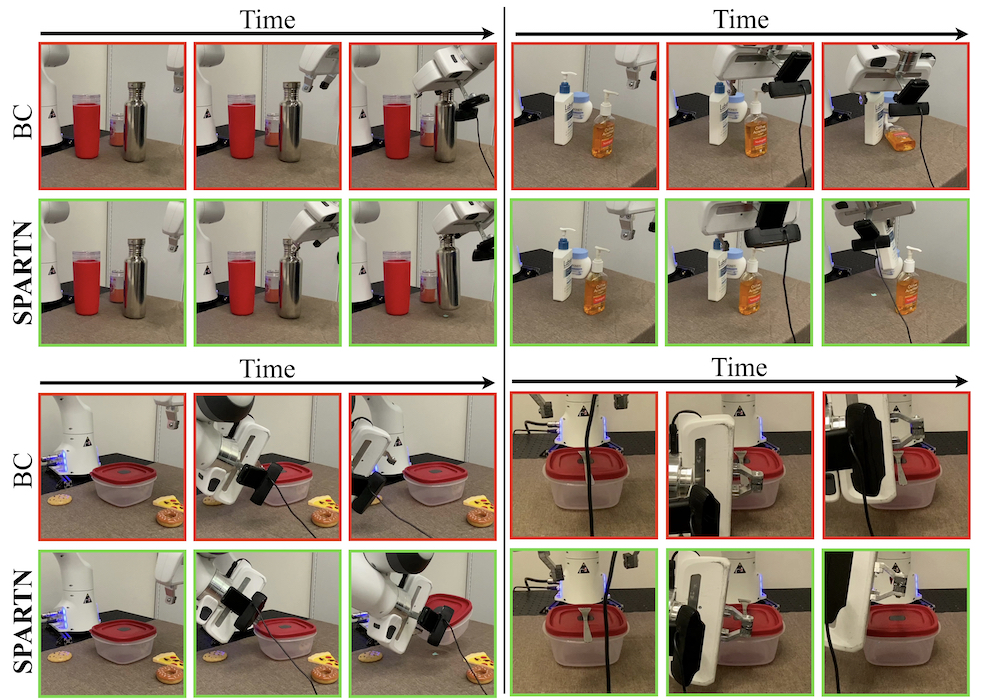}
\definecolor{dark-green}{rgb}{.058,.58,.11}
\definecolor{dark-red}{rgb}{.68,.06,.06}
\vspace{-0.4cm}
\caption{Sample real-world policy rollouts illustrating how \name{} succeeds ({\color{dark-green}green}) in cases where BC fails ({\color{dark-red}red}). \textbf{Top left}: While BC fails to reach the steel bottle, \name{} successfully reaches and grasps it. \textbf{Top right}: While BC collides into the orange bottle (a distractor object), \name{} takes a more rounded path to avoid it before grasping the white Lubriderm bottle. \textbf{Bottom left}: BC fails to recover after a missed grasp, while \name{} successfully reattempts the grasp after failing the first time. \textbf{Bottom right}: \name{} operates with higher precision than BC and successfully completes a difficult fork grasping task.}
\label{fig:rollouts}
\end{figure}

Figure~\ref{fig:rollouts} shows qualitative differences in performance between the BC and \name{} policies. \name{} generally exhibits more reactive behaviors than the baseline policy: it navigates towards the target object better while occasionally avoiding obstacles, executes actions with greater precision, and even reattempts the grasp more successfully after an initial miss. In some cases, the differences are stark: for instance, \name{} may successfully move toward and grasp the target object while the baseline fails to even reach it. We present further analysis of policy rollouts in Appendix~\ref{appendix:additional-qualitative-analysis}, showing how \name{} qualitatively performs better than BC even in cases where both methods fail to grasp the target object. The supplementary videos of real-world policy rollouts illustrate all of these differences, revealing that \name{} induces important reactive closed-loop behaviors that enable the manipulator to successfully execute grasps in the real world.

\section{Conclusion}
We introduce \name{}, which augments eye-in-hand demonstrations with perturbed visual observations and corrective actions. Our augmentation leverages novel-view synthesis, in particular NeRF, to produce these augmentations during \textit{training}-time. \name{} can improve behavior cloning training of robust, real-time, and closed-loop 6-DoF visual control policies. We show that \name{}-trained policies outperform other offline-supervised methods in a simulated 6-DoF grasping generalization benchmark. Our policies can also perform on par with imitation methods that require depth information and online supervision. We verify that \name{} can train policies to grasp a variety of objects in the real world from limited human demonstrations.

Despite its strong performance, \name{} also has some limitations. First, \name{} is limited to tasks with static scenes like grasping: extending to a broader set of manipulation tasks would require effective view synthesis for dynamic scenes. Second, training a neural radiance field for every demonstration before training is computationally expensive, a limitation that may be mitigated through amortized NeRF models~\cite{yu2020pixelnerf,wang2021ibrnet,trevithick2021grf}. Finally, the noise distribution must be tuned for a given platform, e.g. it is tuned separately for the simulated and real experiments. An interesting direction for future work is to use policy actions to generate the noise, which would result in a fully offline variant of DAgger~\cite{ross2011reduction} that uses NeRF as a simulator.

\section{Acknowledgements}
We thank Jimmy Wu, Kaylee Burns, Bohan Wu, and Suraj Nair for technical advice on various aspects of the real robot setup, and Archit Sharma for helpful conceptual discussions. This project was funded by ONR grant N00014-21-1-2685. AZ acknowledges the support of the NSF Graduate Research Fellowship. CF is a fellow of the CIFAR Learning in Machines and Brains Program.

{\small
\bibliographystyle{ieee_fullname}
\bibliography{egbib}
}

\clearpage
\appendix

\section{Full Algorithm}
\begin{algorithm}
\caption{\name{}: Corrective Augmentation via NeRF}
\label{alg:nerf}
\begin{algorithmic}[1]
    \renewcommand{\algorithmicrequire}{\textbf{Input:}}
    \renewcommand{\algorithmicensure}{\textbf{Output:}}
    \renewcommand{\COMMENT}[1]{\textcolor{blue}{\# #1}}
    \REQUIRE $\dset = \{\tau\}_{i=1}^{N}$ - expert demonstrations
    \REQUIRE $\tfm{E}{C}$ - Transform between camera and end-effector
    \REQUIRE $M \in \{0, 1\}^{w \times h}$ - mask for robot gripper
    \ENSURE Augmented transition dataset $\tilde{\dset}$
    \STATE $\tilde{\dset} \gets \{\}$
    \FOR{$\tau = \{(I_k,  \ ^WT^E_k), (^ET^{\hat{E}}_k)\}_{k=1}^K \in \dset$}
        \STATE \COMMENT{EE pose $\rightarrow$ camera pose}
        \STATE $\tfm{W}{C}_k \gets \tfm{W}{E}_k \ \tfm{E}{C}$, for $k=0,\cdots, K$
        \STATE $\{\tfm{V}{C}_k\}_{k=1}^K \gets \textrm{COLMAP}(I_0,\cdots, I_K)$
        \STATE $\beta \gets \textrm{SOLVE}(\textrm{Eq. }\ref{eq:solve-beta})$ \COMMENT{Estimate $\beta$}
        \STATE $\tfm{V}{W}_k = \tfm{V}{C}_k \ (\tfm{W}{C}_k)^{-1}$, for $k=0,\cdots, K$
        \STATE \COMMENT{train NeRF}
        \STATE $F^{\tau}_{\Theta} \gets \text{NeRF}(\{(I_k, \ \tfm{V}{C}_k)\}_{k=1}^K, M)$
        \FOR{$k \in (0, 1, \ldots, K)$}
            \FOR{$i=1:N_{\text{aug}}$}
                \STATE $\varepsilon \gets \text{NoiseDist}(\se)$
                \STATE $^WT^{\tilde{E}}_k \gets \ ^WT^E_k \varepsilon$  \COMMENT{perturbed end-effector pose}
                \STATE $^{\tilde{E}}T^{\hat{E}}_k \gets \varepsilon^{-1} \ ^ET^{\hat{E}}_k $ \COMMENT{corrective relative action}
                \STATE $\tfm{W}{\tilde{C}}_k \gets \ \tfm{W}{\tilde{E}}_k \tfm{E}{C}$ \COMMENT{perturbed camera pose}
                \STATE $\tfm{V}{\tilde{C}}_k \gets \tfm{V}{W}_k \ \tfm{W}{\tilde{C}}_k$
                \STATE $\tilde{I}_k \gets \text{NeRF-Render}(\tfm{V}{\tilde{C}}_k; F^{\tau}_{\Theta})$
                \STATE \COMMENT{splice gripper into rendered image}
                \STATE $\tilde{I}_k \gets \neg M \odot \tilde{I}_k + M \odot I_k$
                \STATE $\tilde{\dset} \gets \tilde{\dset} \cup \{(\tilde{I}_k, \ ^WT^{\tilde{E}}_k), (^{\tilde{E}}T^{\hat{E}}_k)\}$
            \ENDFOR
         \ENDFOR
    \ENDFOR
\end{algorithmic}
\end{algorithm}

\section{Sample Perturbed Observations}
\label{sec:perturbed-samples}
As described in the main text, image observations $\tilde{I}$ for perturbed camera poses are rendered via novel-view synthesis techniques, in particular NeRF. The section ``Example NeRF renders from augmented poses'' in the supplementary website shows videos of the rendering output for both simulation and real-world augmentation.

\section{Simulation Details}
\subsection{Training for Simulation Experiments}
\label{appendix:sim-training}
We train all of our RGB policies (\name{}, DART, and HA) closely following the architectural and optimization hyperparameters of \cite{gaddpg}. The image encoder is a ResNet-18~\cite{he2015deep} pretrained on ImageNet~\cite{russakovsky2015imagenet}, followed by a $3$-layer MLP with $512$ hidden units each and ReLU activations. The MLP outputs the predicted action as a 6-D vector of predicted translations and Euler angles. The final layer has a tanh activation to scale the output between $[-1, 1]$, then each dimension is scaled to match the environment's action bounds.

For behavior cloning, we use the same 3D point-matching loss as \cite{gaddpg}. We optimize the objective using the Adam optimizer with batch size $100$ for $100,000$ steps. We train only the MLP (and not the pre-trained ResNet) for the first $20,000$ steps, since ``freeze-then-train'' fine-tuning techniques have been shown to be more robust~\cite{kumar2022fine}. After $20,000$ steps, the ResNet is unfrozen and the entire network is trained as usual.

\subsection{NeRF Training}
We use Instant-NGP~\cite{mueller2022instant} to train a NeRF for each of the $2{,}500$ demonstration scenes. We train each NeRF for $3{,}500$ steps, which takes $30$ seconds on an NVIDIA GeFORCE RTX 2080 Ti. To reduce total training time, we train the NeRFs in parallel: with $4$ GPUs, we can train all $2{,}500$ NeRFs in $\sim 7$ hours. Because camera calibration is exact in simulation, we use the world-frame camera poses given by calibration instead of COLMAP to train NeRF.

\section{Real-World Details}
\subsection{Policy Training}
\label{sec:real-world-policy-training}
For both BC and \name{}, the image encoder for each policy is a four-layer convolutional network followed by two feedforward layers, and uses batch normalization and ReLU activation functions. The image embedding is fed into a three-layer ReLU MLP policy head, which outputs the predicted action. We train the policy from scratch with mean squared error, using Adam with learning rate $5\times 10^{-4}$ and a batch size of $64$. No image augmentations like random crop are applied.

\subsection{COLMAP and NeRF Training}
\label{appendix:colmap-nerf}

In the real world, we train our NeRF using camera poses $\{\Htfm{V}{C}_k\}_{k=1}^K$ estimated by COLMAP~\cite{schoenberger2016sfm,schoenberger2016mvs}. From the camera calibration, we also have pose transforms $\{\tfm{W}{C}_k\}_{k=1}^K$ which are too noisy to train a good NeRF from, but will allow us to convert between world and COLMAP frames. Conversion is necessary because we want to render from perturbed camera poses $\tfm{W}{\tilde{C}}$, but we must call NeRF-Render using the transform $\Htfm{V}{\tilde{C}}$ (COLMAP's reference frame) instead. From Eq.~\ref{eq:colmap}, we can do this conversion if we have the transform $\tfm{V}{W}$ and the scale factor $\beta$ which distinguishes $T$ and $H$.

We can estimate both quantities using the $(\tfm{W}{C}_k, \Htfm{V}{C}_k)$ pairs in each demonstration. Recalling that $\tfm{a}{b} = (\tfm{b}{a})^{-1}$, we have for any $j\neq k$:
\begin{equation}
    \label{eq:solve-beta}
    \beta \ \T{\tfm{C}{W}_j \ \tfm{W}{C}_k} = \T{\Htfm{C}{V}_j \ \Htfm{V}{C}_k}
\end{equation}
This leads to an overconstrained linear system in one unknown $\beta$, which can be solved with linear regression.
After solving for $\beta$, we can rescale from $H \rightarrow T$:
\begin{equation}
    \tfm{V}{C}_k := \left(\R{\Htfm{C}{V}_k}, (1 / \beta) \ \T{\Htfm{C}{V}_k}\right)
\end{equation}
Now solving for $\tfm{V}{W}$ is simple:
\begin{equation}
    \tfm{V}{W}_k = \tfm{V}{C}_k (\tfm{W}{C}_k)^{-1}.
\end{equation}
In theory, $\tfm{V}{W}$ is constant for all $k$, but in practice there is noise from both COLMAP and the real-world pose estimates, so we simply compute $\tfm{V}{W}_k$ separately for each $k$.

\subsection{Additional Qualitative Analysis}
\label{appendix:additional-qualitative-analysis}

\begin{figure}
\includegraphics[width=0.48\textwidth]{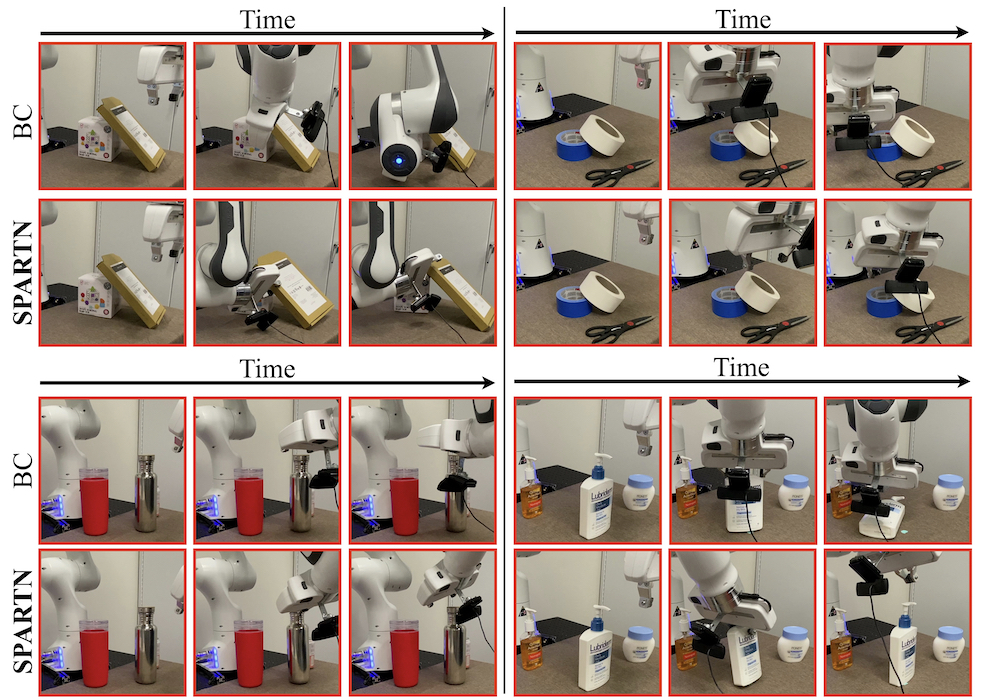}
\caption{Sample real-world policy rollouts illustrating cases where both BC and \name{} policies fail. \textbf{Top left}: BC fails to reach the thin box, while \name{} successfully reaches it and nearly completes the grasp. \textbf{Top right}: While BC completely misses the white tape roll, \name{} makes contact but fails to grasp it. \textbf{Bottom left}: BC fails to reach the steel water bottle, while \name{} contacts it but fails to grasp it. \textbf{Bottom right}: BC reaches the Lubriderm bottle over but knocks it over before grasping it, while \name{} nearly completes the grasp as the bottle barely slips away from the robot's fingers.}
\label{fig:failure-rollouts}
\end{figure}

Figure \ref{fig:failure-rollouts} shows sample evaluation trials where both SPARTN and BC policies fail, given the same initial object and end-effector configurations. Even in these failure cases, SPARTN qualitatively performs better than BC, nearly grasping the object in some cases, and dropping the object soon after grasping in other cases. See the videos on the supplemental website for additional qualitative results.

\subsection{Real-World Environments}
\label{sec:real-world-environments}

The real-world environments shown in Figure \ref{fig:environments} each include a single target object to grasp among other distractor objects. The initial positions of all objects and the robotic end-effector are randomized while collecting expert demonstrations and during test time. The initial orientations of the objects are also randomized such that all policies must perform end-effector orientation control to complete the task. To make a fair comparison, we evaluate different policies given the same set of initial configurations (as shown in Figure \ref{fig:rollouts}, Figure \ref{fig:failure-rollouts}, and the videos on the supplemental website); we do this by resetting the robot and environment to the same conditions and alternating the rollouts of different policies before moving on to the next test configuration. Rollouts are considered successful if the robot grasps the target object and lifts it steadily into the air. We describe the details of each environment below:

\begin{itemize}
  \item \textbf{Banana}: The environment includes an artificial banana resting on a white ceramic plate, a red cup, an artificial avocado, an artificial tomato, and a plastic fork and knife resting on a blue cloth.
  \item \textbf{Thin box}: The environment includes a thin brown box resting on a thicker white toy box.
  \item \textbf{Steel water bottle}: The environment includes a steel water bottle, a red water bottle, and an aluminum can. The steel water bottle is highly reflective.
  \item \textbf{Wine glass}: The environment includes a wine glass, red water bottle, and blue water bottle all placed upside-down and leaning against a brown box. The wine glass is transparent.
  \item \textbf{Lubriderm bottle}: The environment includes a white Lubriderm bottle, an orange Neutrogena bottle, and a small white container of moisturizer.
  \item \textbf{White tape roll}: The environment includes a roll of white tape resting on a roll of blue tape, as well as a pair of black scissors. The rolls of tape are radially symmetric.
  \item \textbf{Tupperware}: The environment includes a plastic tupperware container with a red lid, as well as three artificial food items: a donut, a cookie, and a slice of pizza.
  \item \textbf{Fork}: The environment includes a fork resting on top of a plastic tupperware container. The fork's location and orientation relative to the surface of the tupperware container is fixed, but the position of the entire assembly is randomized.
\end{itemize}

\end{document}